\documentclass[preprint,12pt]{elsarticle}

\usepackage{amsmath}
\usepackage{amssymb}
\usepackage{bm}
\usepackage{algorithm}
\usepackage{algorithmicx}
\usepackage{algpseudocode}
\usepackage{diagbox}
\usepackage{graphicx}
\usepackage{url}
\usepackage{csquotes}
\usepackage{mathrsfs}
\usepackage{amsthm}
\usepackage{mleftright}

\newcommand{\positive}{{\mathrm{P}}}
\newcommand{\negative}{{\mathrm{N}}}
\newcommand{\unlabeled}{{\mathrm{U}}}

\newcommand{\expts}[2]{\ensuremath{\mathbb{E}_{#1}\left[{#2}\right]}}
\newcommand{\argmax}{\mathop{\rm arg~max}\limits}
\newcommand{\argmin}{\mathop{\rm arg~min}\limits}

\renewcommand{\hat}{\widehat}
\renewcommand{\tilde}{\widetilde}
\def\D{\mathrm{d}}
\def\numx#1e#2{{#1}\mathrm{e}{#2}}
\newtheorem{theorem}{Theorem}

\renewcommand{\mkbegdispquote}[2]{\itshape}

\begin{document}

% ----- metadata ----- %
\begin{frontmatter}
\title{Convex Formulation of Multiple Instance Learning from Positive and Unlabeled Bags}

\author[1,3]{Han Bao\corref{cor}}
\ead{tsutsumi@ms.k.u-tokyo.ac.jp}

\author[2,3]{Tomoya Sakai}
\ead{sakai@ms.k.u-tokyo.ac.jp}

\author[2,1,3]{Issei Sato}
\ead{sato@k.u-tokyo.ac.jp}

\author[3,4,2,1]{Masashi Sugiyama}
\ead{sugi@k.u-tokyo.ac.jp}

\cortext[cor]{Corresponding author}

\address[1]{Department of Computer Science, The University of Tokyo}
\address[2]{Department of Complexity Science and Engineering, The University of Tokyo}
\address[3]{Center for Advanced Intelligence Project, RIKEN}
\address[4]{International Research Center for Neurointelligence, The University of Tokyo}

% ----- abstract ----- %
\begin{abstract}
Multiple instance learning (MIL) is a variation of traditional supervised learning problems where data (referred to as bags) are composed of sub-elements (referred to as instances) and only bag labels are available.
MIL has a variety of applications such as content-based image retrieval, text categorization, and medical diagnosis.
Most of the previous work for MIL assume that training bags are fully labeled.
However, it is often difficult to obtain an enough number of labeled bags in practical situations, while many unlabeled bags are available.
A learning framework called PU classification (positive and unlabeled classification) can address this problem.
In this paper, we propose a convex PU classification method to solve an MIL problem.
We experimentally show that the proposed method achieves better performance with significantly lower computation costs than an existing method for PU-MIL.

\end{abstract}


\end{frontmatter}

% ----- introduction ----- %

\section{Introduction}
\label{sec:introduction}

% 背景
Multiple instance learning (MIL)~\cite{Dietterich:1997} is a learning problem with {\it bags} and {\it instances}.
Instances are the same as ordinary feature vectors, while bags are sets of instances.
The numbers of instances in different bags varies.
Bag labels are defined as follows.
\begin{itemize}
    \item If a bag contains {\it at least one positive instance}, then its label is positive.
    \item If a bag contains {\it no positive instances}, then its label is negative.
\end{itemize}
This is the basic setup of MIL.
The goal of MIL is to predict labels of test bags.
MIL is more difficult than ordinary classification problems because instance labels are unavailable.

MIL was originated from molecule/graph data~\cite{Dietterich:1997}, where ray-based representation is used to describe molecule shapes.
Later, MIL has been considered as a graph-based learning problem~\cite{Wu:2013:ICDM,Wu:2014:SDM,Wu:2017:NNLS,Wu:2017:TC,Wu:2018:TKDE}.
In fact, MIL is applicable to a wide range of real-world problems such as molecule behavior prediction~\cite{Lindsay:1980}, drug activity prediction~\cite{Dietterich:1997}, domain theory~\cite{Dietterich:1988}, content-based image retrieval~\cite{Maron:1998b,Li:2015,Wu:2015}, visual tracking~\cite{Babenko:2009}, object detection~\cite{Pandey:2011,Kanezaki:2011}, text categorization~\cite{Andrews:2002}, and medical diagnosis~\cite{Dundar:2008,Tong:2014}.

So far, a lot of approaches for MIL have been developed~\cite{Dietterich:1997,Maron:1998,Zhang:2001,Andrews:2002,Gartner:2002,Wang:2000},
which are classified into two groups in general.
\begin{enumerate}
\item
Methods in the first group are based on generative modeling, including the diverse density~\cite{Maron:1998} and its extension, the expectation-maximization diverse density (EM-DD)~\cite{Zhang:2001}.
These methods find out an instance close to instances in training positive bags and far from instances in training negative bags, which is referred to as a concept point.
This process is carried out by gradient-based search from every training instance, which is computationally inefficient. \\

\item
Methods in the second group are based on discriminative modeling.
The multiple-instance support vector machine (MI-SVM)~\cite{Andrews:2002} is an approach based on SVMs.
Empirical evaluation shows that MI-SVM performs well, but its optimization problem is non-convex and finding a solution is computationally expensive.
The key-instance support vector machine (KI-SVM)~\cite{Li:2009:ECML} reformulates the optimization problem of MI-SVM as mixed-integer programming, which is still hard to optimize.
G\"{a}rtner et al.~\cite{Gartner:2002} introduced set kernels (a.k.a.~multiple instance kernels), which are extensions of the standard kernel functions to MIL.
The set kernels can be used to construct a standard SVM classifier, which performs well in experiments.
The optimization problem in this training procedure is convex and the global solution can be obtained efficiently.
\end{enumerate}

%These standard approaches to MIL assume that training bags are fully labeled.
%However, it is often difficult to obtain an enough amount of labeled data while many unlabeled data are often available in practical situations such as outlier detection~\cite{Hido:2008}.
%Rahmani and Goldman~\cite{Rahmani:2006} proposed semi-supervised MIL based on the label propagation, which needs a particular smoothness assumption on labels, but this assumption does not always hold for real datasets.

% 提案法, 研究目的, 動機
In this work, we propose a novel method to construct multiple instance classifiers only from positive and unlabeled bags, while the above standard approaches to MIL assume that training bags are fully labeled.
This problem is called PU-MIL.
For example, PU-MIL is applicable to the following situations.
\begin{itemize}
  \item
  The situation where it is difficult to obtain an enough amount of labeled data due to the significant labeling costs, such as outlier detection based on supervised classification,
  where it is often difficult to label all outlier samples.
  On the other hand, in PU-MIL, we need to label only some of outlier samples and the rest can be regarded as unlabeled.

  \item
  The situation where the true negative labels are essentially unavailable, such as bioinformatics and cheminformatics.
  MIL setting commonly appears in these natural science fields~\cite{Dietterich:1997}.
  In natural science, experiments are often designed to observe some phenomena (detect positives), not designed to deny the existence of the phenomena.
  Thus even if we did not observe the phenomenon, it might not be appropriate to say ``the phenomenon did not occur.''
  In other words, there might be {\it false negatives}.
  PU setting plays an important role in this kind of situations.
\end{itemize}

%In this work, we propose a novel method to construct multiple instance classifiers only from positive and unlabeled bags.
%This problem is called PU-MIL.
%PU-MIL is applicable to, e.g., music album recommendation, where the goal is to predict music albums that consumers want to purchase from their purchase history.
%Here each album contains some songs, which can be regarded as multiple instances (each album is a bag and its songs are instances).
%The positive label means that the album attracts the consumer, while the negative label means that the album does not attract the consumer.
%However, the labeling process for the training data is problematic: albums purchased by the consumer in the past can be positively labeled, but albums not purchased cannot be negatively labeled straightforwardly.
%The fact that the consumer did not purchase albums does not directly mean that they are not attracting.
%In such a situation, albums not purchased can be regarded as unlabeled bags.

Our contribution in this paper is to propose a novel PU-MIL method based on empirical risk minimization~\cite{Plessis:2015}.
The proposed method formulates an optimization problem as a convex optimization problem together with a linear-in-parameter model, and the global optimal solution can be computed efficiently.
To the best of our knowledge, this is the first convex PU-MIL method (see Table \ref{tab:MIL-methods}).
Through experiments, we demonstrate that the proposed method combined with the minimax kernel~\cite{Andrews:2002} compares favorably with an existing method.

\begin{table}[t]
\begin{center}
    \caption{Comparison of existing and proposed discriminative methods for MIL.}
    \begin{tabular}{|l||c|c|} \hline
    \backslashbox{Learning from}{Convexity} & Convex & Non-convex \\ \hline \hline
    \shortstack{Positive and Negative \\ \\ \\ \\ \\ \\ $\;$}
        & \shortstack{ \\ \\
            set kernels~\cite{Gartner:2002} \\
            sMIL~\cite{Bunescu:2007} \\
            KI-SVM~\cite{Li:2009:ECML} \\
            miGraph~\cite{Zhou:2009} \\
            $\;$
        }
        & \shortstack{ \\ \\
            MI-SVM~\cite{Andrews:2002} \\
            MissSVM~\cite{Zhou:2007} \\
            soft-bag SVM~\cite{Li:2015} \\
            dMIL~\cite{Wu:2015} \\
            $\;$
        }
    \\ \hline
    \shortstack{Positive and Unlabeled \\ $\;$}
        & \shortstack{ \\ \\
            {\bf PU-SKC} (Sect.~\ref{sec:method}) \\ $\;$
        }
        & \shortstack{ \\ \\
            puMIL~\cite{Wu:2014} \\ $\;$
        }
    \\ \hline
    \end{tabular}
    \label{tab:MIL-methods}
\end{center}
\end{table}

The rest of this paper is structured as follows.
In Sect.~\ref{sec:related-work}, we review existing methods for PU classification~\cite{Plessis:2014,Plessis:2015} and MIL~\cite{Gartner:2002}, on which our proposed method is based.
In Sect.~\ref{sec:method}, we explain the formulation and optimization algorithm of our proposed method, called the {\it positive and unlabeled set kernel classifier} (PU-SKC).
In Sect.~\ref{sec:experiments}, we experimentally compare the performance of the proposed method (PU-SKC) with an existing method (puMIL)~\cite{Wu:2014}.
Finally, we conclude this work in Sect.~\ref{sec:conclusion}.

% ----- related work ----- %

\section{Problem Formulation and Related Work}
\label{sec:related-work}

\begin{figure}
  \begin{center}
    \includegraphics[width=12cm]{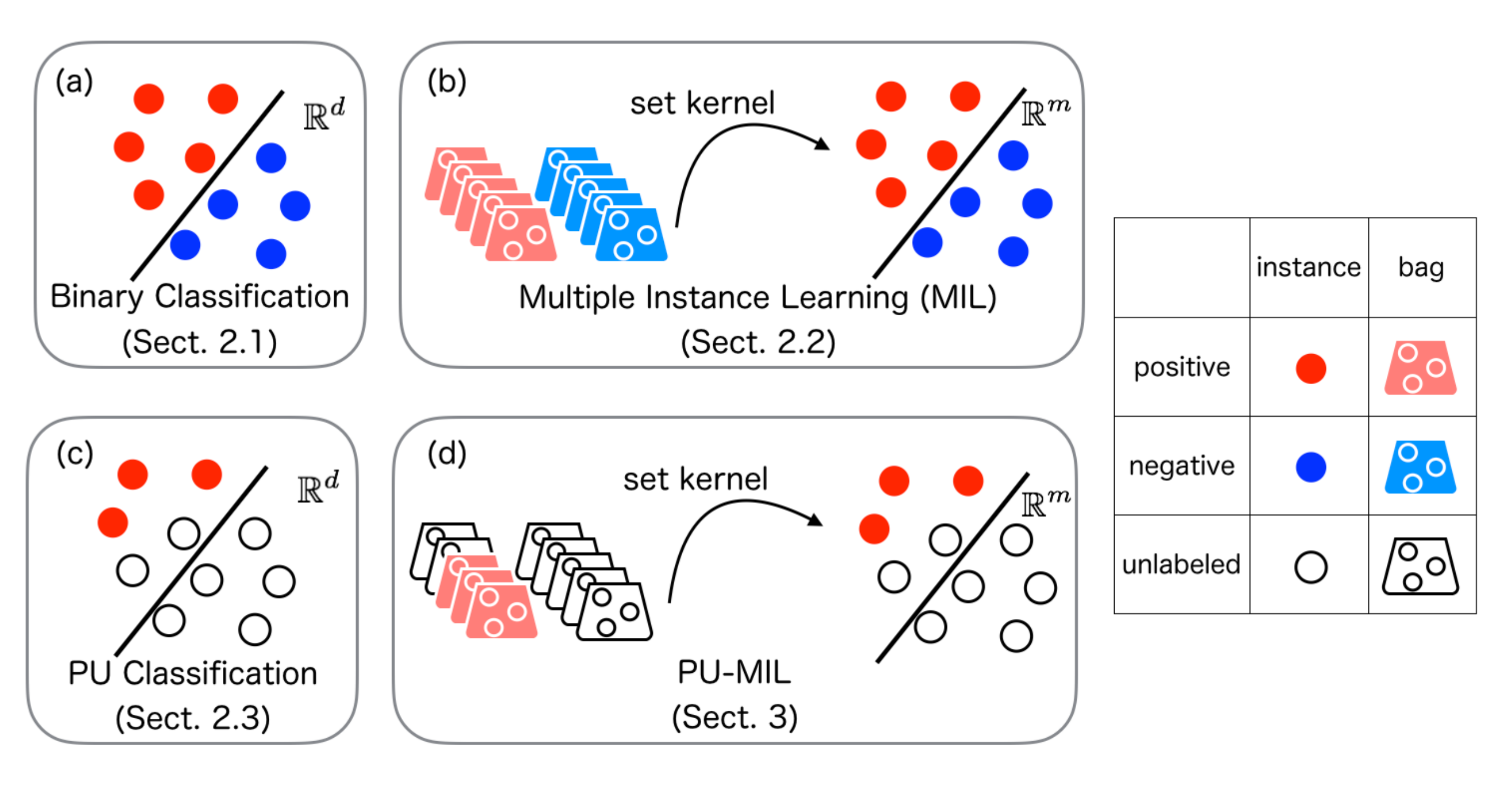}
    \caption{A schematic of problems. In this work, we consider (d) multiple instance learning from positive and unlabeled bags (PU-MIL).}
    \label{fig:schematic}
  \end{center}
\end{figure}

In this section, we formulate the problems we discuss in this paper (see Fig. \ref{fig:schematic}) and review related work.

\subsection{Ordinary Binary Classification}
\label{sec:binary-classification}

Let $\bm{x} \in \mathbb{R}^d$ be a $d$-dimensional feature vector and $y \in \{+1, -1\}$ be its corresponding class label.
In the ordinary binary classification problem, we construct a binary classifier
\begin{align}
f(\bm{x}) & = \mathrm{sign}(g(\bm{x})) \in \{+1, -1\}, \nonumber \\
g & : \mathbb{R}^d \to \mathbb{R}, \label{eq:binary-classifier}
\end{align}
from an i.i.d.~training dataset $\mathcal{D} = \{(\bm{x}_i, y_i)\}_{i=1}^N$, where $N$ is the number of training samples.
Here we use a linear-in-parameter model for $g$:
\[
g(\bm{x}) = \bm{\alpha}^\top\bm{\phi}(\bm{x}) + \beta ,
\]
where $\top$ denotes the transpose, $\bm{\alpha} \in \mathbb{R}^m$ is an $m$-dimensional parameter vector, $\beta \in \mathbb{R}$ is a bias parameter, and $\bm{\phi} : \mathbb{R}^d \to \mathbb{R}^m$ is a vector of basis functions.
The support vector machine (SVM)~\cite{Cortes:1995} is one of the most standard methods for training a binary classifier.
The optimization problem of SVM is given as follows:
\begin{align}
\min_{\bm{\alpha} \in \mathbb{R}^m, \beta \in \mathbb{R}} \quad
& \frac{1}{2}||\bm{\alpha}||^2
  + C \sum_{i=1}^N \max \left\{ 0, 1 - y_i \left( \bm{\alpha}^\top \bm{\phi}(\bm{x}_i) + \beta \right) \right\} \label{eq:svm} ,
\end{align}
where $C > 0$ is a penalty parameter.
This problem can be reformulated as a quadratic program (QP), which can be solved efficiently.

\subsection{Multiple Instance Learning}
We formulate the problem of multiple instance learning (MIL) and review an existing method.

\subsubsection{Formulation}

Hereafter $\mathcal{P}(A)$ denotes the power set\footnote{
  The power set of $A$ is a set of all subsets of $A$, including $\emptyset$ and $A$ itself.
  In the MIL setting, bags belong to $\mathcal{P}(\mathbb{R}^d)$, i.e., bags are composed of some elements in $\mathbb{R}^d$.
} of $A$.
Let $X = \{\bm{x}_i | \bm{x}_i \in \mathbb{R}^d\}_{i=1}^n \in \mathcal{P}(\mathbb{R}^d) \setminus \emptyset$ be a bag containing $n$ instances whose dimensions are $d$, and $Y \in \{+1, -1\}$ be a bag label corresponding to $X$.
The problem is to construct a binary classifier:
\begin{align}
f(X) &= \mathrm{sign}(g(X)) \in \{+1, -1\} , \nonumber \\
g &: \mathcal{P}(\mathbb{R}^d) \setminus \emptyset \to \mathbb{R} , \label{eq:bag-classifier}
\end{align}
from an i.i.d.~fully-labeled training dataset $\mathcal{D} = \{(X_b, Y_b)\}_{b=1}^{N}$,
where $N$ denotes the number of bags in $\mathcal{D}$.

\subsubsection{Multiple Instance Kernels}
\label{sec:mil}

G\"{a}rtner et al.~\cite{Gartner:2002} proposed set kernels (multiple instance kernels),
which map bags (sets of instances) to a feature space.
A type of the set kernels, called the statistic kernel $\widetilde{k}$, is defined as follows:
\[
\widetilde{k}(X, X') := k(\bm{s}(X), \bm{s}(X')) ,
\]
where $k$ is an arbitrary kernel function such as the Gaussian kernel, and $\bm{s}$ is called a statistic.
For example, the following minimax statistics is a typical choice:
\begin{align}
\bm{s}_{\mathrm{minimax}}(X) & := \left[ \min_{\bm{x} \in X} x^{(1)}, \; \ldots, \; \min_{\bm{x} \in X} x^{(d)}, \; \max_{\bm{x} \in X} x^{(1)}, \; \ldots, \; \max_{\bm{x} \in X} x^{(d)} \right]^\top , \label{eq:minimax-statistics}
\end{align}
where $x^{(i)}$ is the $i$-th element of an instance $\bm{x}$ in the bag $X$.
G\"{a}rtner et al.~\cite{Gartner:2002} experimentally demonstrated that the statistic kernel with the minimax statistics~\eqref{eq:minimax-statistics} for $\bm{s}$ and the polynomial kernel for $k$ shows good performance:
\begin{align}
\widetilde{k}_\mathrm{minimax}(X, X') & := (\bm{s}_{\mathrm{minimax}}(X)^\top \bm{s}_{\mathrm{minimax}}(X') + 1)^\rho \label{eq:set-kernel} ,
\end{align}
where $\rho$ is a positive integer.
The statistic kernel~\eqref{eq:set-kernel} is referred to as the minimax kernel.
We can then construct the following set kernel classifier $g$:
\begin{align}
g(X) & = \bm{\alpha}^\top\bm{\phi}(X) + \beta \label{eq:mi-classifier} , \\
\bm{\phi}(X) & = \left[\begin{array}{c}
\widetilde{k}_\mathrm{minimax}(X, C_1) \\
\vdots \\
\widetilde{k}_\mathrm{minimax}(X, C_M)
\end{array}\right] ,
\nonumber
\end{align}
where $C_1, \ldots, C_M$ are kernel centers and $M$ is the number of kernel centers.
We can obtain the MIL classifier by using SVM~\eqref{eq:svm} to train the classifier~\eqref{eq:mi-classifier}.

\subsection{Learning from Positive and Unlabeled Data}
\label{sec:pu-learning}

We formulate a binary classification problem from positive and unlabeled instances and review existing methods.

\subsubsection{Formulation}

We assume that positive samples $\{\bm{x}_i^\positive\}_{i=1}^{N_\positive}$ and unlabeled samples $\{\bm{x}_i^\unlabeled\}_{i=1}^{N_\unlabeled}$ are generated as follows:
\begin{align}
\{\bm{x}_i^\positive\}_{i=1}^{N_\positive}   \stackrel{\mathrm{i.i.d.}}{\sim} &\; p_+(\bm{x}) := p(\bm{x}|y=+1), \nonumber \\
\{\bm{x}_i^\unlabeled\}_{i=1}^{N_\unlabeled} \stackrel{\mathrm{i.i.d.}}{\sim} &\; p(\bm{x}) := \pi p(\bm{x}|y=+1) + (1-\pi) p(\bm{x}|y=-1) \nonumber \\
& \hspace{24pt} = \pi p_+(\bm{x}) + (1 - \pi) p_-(\bm{x}), \label{eq:unlabeled-sample-distribution}
\end{align}
where $\pi := p(y = +1)$ is called the class prior.
Our objective is to construct the binary classifier~\eqref{eq:binary-classifier} only from positive and unlabeled samples.

\subsubsection{Learning Instance-Level Classifiers from Positive and Unlabeled Data}

du Plessis et al.~\cite{Plessis:2014,Plessis:2015} proposed methods based on empirical risk minimization to learn only from positive and unlabeled samples.
In the ordinary binary classification setting, an optimal classifier $g^*$ minimizes the following misclassification rate:
\begin{align}
R_{0\mbox{-}1}(g) = \pi \expts{\positive}{\ell_{0\mbox{-}1}(g(\bm{x}))} + (1-\pi)\expts{\negative}{\ell_{0\mbox{-}1}(-g(\bm{x}))} , \label{eq:zero-one-risk}
\end{align}
where $\expts{\positive}{\cdot}$ and $\expts{\negative}{\cdot}$ denote the expectations over $p_+(\bm{x})$ and $p_-(\bm{x})$ respectively and $\ell_{0\mbox{-}1}$ denotes the zero-one loss:
\begin{align}
\ell_{0\mbox{-}1}(z) =
\begin{cases}
0 & \mbox{if $z \ge 0$,} \\
1 & \mbox{otherwise.}
\end{cases}
\nonumber
\end{align}
In practice, the misclassification rate~\eqref{eq:zero-one-risk} is difficult to optimize because the subgradient of $\ell_{0\mbox{-}1}$ is always $0$ except at $z = 0$.
For this reason, we usually use a {\it surrogate} loss function\footnote{
  For example, the hinge loss $\ell_{\mathrm{H}}(z) = \max(-z, 0)$ and the ramp loss $\ell_{\mathrm{R}}(z) = \frac{1}{2}\max(0, \min(2, 1-z))$ are commonly used~\cite{Vapnik:1995}.
}.
Then the risk function $R$ with the surrogate loss function $\ell$ is written as
\begin{align}
R(g) := \pi \expts{\positive}{\ell(g(\bm{x}))} + (1 - \pi) \expts{\negative}{\ell(-g(\bm{x}))} . \label{eq:PN-risk}
\end{align}
Since negative samples are not available in the PU classification setup,
let us consider expressing the risk~\eqref{eq:PN-risk} without $\expts{\negative}{\cdot}$.
By the definition of the unlabeled sample distribution~\eqref{eq:unlabeled-sample-distribution}, the following equation holds:
\begin{align*}
(1 - \pi) \expts{\negative}{\ell(-g(\bm{x}))} = \expts{\unlabeled}{\ell(-g(\bm{x}))} - \pi \expts{\positive}{\ell(-g(\bm{x}))} .
\end{align*}
Substituting this into the risk~\eqref{eq:PN-risk}, we obtain
\begin{align}
R(g) = \pi \expts{\positive}{\ell(g(\bm{x})) - \ell(-g(\bm{x}))} + \expts{\unlabeled}{\ell(-g(\bm{x}))} , \label{eq:PU-risk}
\end{align}
where $\expts{\unlabeled}{\cdot}$ denotes the expectation over $p(\bm{x})$.
If the surrogate loss function $\ell$ satisfies
\begin{align}
\ell(z) - \ell(-z) = -z \label{eq:linear-composite-loss} ,
\end{align}
the risk~\eqref{eq:PU-risk} can be written as
\begin{align}
R(g) = \pi \expts{\positive}{-g(\bm{x})} + \expts{\unlabeled}{\ell(-g(\bm{x}))} . \label{eq:PU-convex-obj}
\end{align}
The risk~\eqref{eq:PU-convex-obj} is convex if the surrogate loss function $\ell$ is convex.
Convex loss functions such as the squared loss $\ell_\mathrm{S}$, the logistic loss $\ell_\mathrm{LL}$, and the double hinge loss $\ell_\mathrm{DH}$ satisfy the condition~\eqref{eq:linear-composite-loss}:
\begin{align}
\ell_\mathrm{S}(z)  &= \frac{1}{4}(z - 1)^2, \nonumber \\
\ell_\mathrm{LL}(z) &= \log(1 + \exp(-z)), \nonumber \\
\ell_\mathrm{DH}(z) &= \max\left(-z, \max\left(0, \frac{1-z}{2}\right)\right) . \label{eq:dh-loss}
\end{align}

We use the risk~\eqref{eq:PU-convex-obj} to obtain a convex formulation of PU-MIL in Sect.~\ref{sec:method}.

% ----- method ----- %

\section{Positive and Unlabeled Set Kernel Classifier}
\label{sec:method}

In this section, we propose a convex method for PU-MIL, named the PU-SKC (positive and unlabeled set kernel classifier).

\subsection{Multiple Instance Learning from Positive and Unlabeled Bags}

We formulate the problem of multiple instance learning from positive and unlabeled bags (PU-MIL).
The purpose of PU-MIL is to construct the bag-level classifier~\eqref{eq:bag-classifier}
from a positively labeled training dataset $\mathcal{D}_\positive = \{(X_b^\positive, Y_b = +1)\}_{b=1}^{N_\positive}$ and an unlabeled training dataset $\mathcal{D}_\unlabeled = \{X_{b'}^\unlabeled\}_{{b'}=1}^{N_\unlabeled}$,
where $N_\positive$ and $N_\unlabeled$ denote the number of positive bags in $\mathcal{D}_\positive$ and the number of unlabeled bags in $\mathcal{D}_\unlabeled$, respectively.
We assume that $X_1^\positive, \dots, X_{N_\positive}^\positive \stackrel{\mathrm{i.i.d.}}{\sim} p(X|Y=+1)$ and $X_1^\unlabeled, \dots, X_{N_\unlabeled}^\unlabeled \stackrel{\mathrm{i.i.d.}}{\sim} p(X)$.

\subsection{Formulation}

As we mentioned in Sect.~\ref{sec:pu-learning}, du Plessis et al.~\cite{Plessis:2014,Plessis:2015} formulated the PU classification problem in the empirical risk minimization framework.
If we use a loss function $l(z)$ such that $l(z) - l(-z) = -z$, we have the following objective function:
\begin{align}
J(g) = \pi \expts{\positive}{-g(X)} + \expts{\unlabeled}{\ell(-g(X))} \label{eq:convex-risk} .
\end{align}
Here we use a linear-in-parameter model with the set kernel function as a classifier:
\begin{align}
g(X) = \bm{\alpha}^\top\bm{\phi}(X) + \beta , \label{eq:bag-clf}
\end{align}
where $\bm{\phi}$ is a vector of basis functions:
\begin{align}
\bm{\phi}(X) = \left[\begin{array}{c}
\widetilde{k}_\mathrm{minimax}(X, X^\positive_1) \\
\vdots \\
\widetilde{k}_\mathrm{minimax}(X, X^\positive_{N_\positive}) \\
\widetilde{k}_\mathrm{minimax}(X, X^\unlabeled_1) \\
\vdots \\
\widetilde{k}_\mathrm{minimax}(X, X^\unlabeled_{N_\unlabeled})
\end{array}\right] . \label{eq:bag-basis}
\end{align}
As with the standard binary classification, we predict a given bag as positive if $g(X) \ge 0$, and as negative if $g(X) < 0$.

The risk~\eqref{eq:convex-risk} together with the bag-level classifier~\eqref{eq:bag-clf} and the $l_2$ regularizer induces the following objective function to be minimized:
\begin{align}
\widehat{J}(\bm{\alpha},b)
& = \pi \cdot \frac{1}{N_\positive} \sum_{b=1}^{N_\positive} \left( - \bm{\alpha}^\top\bm{\phi}(X^\positive_b) - \beta \right) \nonumber \\
  & + \frac{1}{N_\unlabeled} \sum_{{b'}=1}^{N_\unlabeled} \ell_\mathrm{DH}\left(-\bm{\alpha}^\top\bm{\phi}(X^\unlabeled_{b'}) - \beta\right)
    + \frac{\lambda}{2}\bm{\alpha}^\top\bm{\alpha} , \label{eq:pu-skc-obj}
\end{align}
where $\lambda \ge 0$ is the regularization parameter.
Here we use the double hinge loss $\ell_\mathrm{DH}$~\eqref{eq:dh-loss} because du Plessis et al.~\cite{Plessis:2015} reported that it achieved the best performance in the ordinary PU classification setting.
Note that $\pi$ is the bag-level class prior, i.e., $\pi = p(Y = +1)$, which must be estimated from the training data.
We explain how to estimate it in Sect.~\ref{sec:bag-class-prior-estimation}.

The problem of minimizing~\eqref{eq:pu-skc-obj} can be rewritten in the form of a quadratic program by using slack variables $\bm{\xi}$ as
\begin{align}
\min_{\bm{\alpha},\beta,\bm{\xi}} \quad &
  - \frac{\pi}{N_\positive}\bm{1}^\top\Phi_\positive\bm{\alpha}
  - \pi \beta
  + \frac{1}{N_\unlabeled}\bm{1}^\top\bm{\xi}
  + \frac{\lambda}{2}\bm{\alpha}^\top\bm{\alpha} \label{eq:puskc-qp} \\
\mbox{s.t.} \quad
& \bm{\xi} \geq \bm{0} , \nonumber \\
& \bm{\xi} \geq \frac{1}{2}\bm{1} + \frac{1}{2}\Phi_\unlabeled\bm{\alpha} + \frac{1}{2}\beta\bm{1} , \nonumber \\
& \bm{\xi} \geq \Phi_\unlabeled\bm{\alpha} + \beta\bm{1} , \nonumber
\end{align}
where $\geq$ for vectors denotes the element-wise inequality, and $\bm{0}$, $\bm{1}$ denote the all-zero and all-one vectors, respectively.
Matrices $\Phi_\positive$ and $\Phi_\unlabeled$ are defined as follows:
\[
\Phi_\positive = \left[\begin{array}{c}
\bm{\phi}^\top(X^\positive_1) \\
\vdots \\
\bm{\phi}^\top(X^\positive_{N_\positive})
\end{array}\right], \quad
\Phi_\unlabeled = \left[\begin{array}{c}
\bm{\phi}^\top(X^\unlabeled_1) \\
\vdots \\
\bm{\phi}^\top(X^\unlabeled_{N_\unlabeled})
\end{array}\right] .
\]

Note that our proposed method is independent of kernel choices.
Other set kernels, such as the conformal kernels~\cite{Blaschko:2006} and affinity propagation clustering-based feature representation~\cite{Li:2014:VCIR},
can also be used instead of the polynomial minimax kernel,
depending on domain-specific knowledge.

\subsection{Bag-Level Class Prior Estimation}
\label{sec:bag-class-prior-estimation}

A bag-level class prior estimation algorithm can be obtained by a simple extension of the instance-level version explained in \ref{sec:class-prior-estimation}.
The difference is basis functions used for estimating the class prior.
We use the polynomial minimax kernel~\eqref{eq:set-kernel} to obtain the bag-level basis functions~\eqref{eq:bag-basis}.
Then the bag-level class prior $\pi$ can be estimated similarly:
\begin{align*}
  \pi = \left[ 2\hat{\bm{h}}^\top\hat{G}^{-1}\hat{\bm{h}} - \hat{\bm{h}}^\top\hat{G}^{-1}\hat{H}\hat{G}^{-1}\hat{\bm{h}} \right]^{-1},
\end{align*}
where
\begin{align*}
  \hat{H} = \frac{1}{N_\positive} \sum_{b=1}^{N_\positive} \bm{\phi}(X_b^\positive)\bm{\phi}(X_b^\positive)^\top, \quad
  \hat{\bm{h}} = \frac{1}{N_\unlabeled} \sum_{b'=1}^{N_\unlabeled} \bm{\phi}(X_{b'}^\unlabeled), \quad
  \hat{G} = \hat{H} + \eta I,
\end{align*}
and $\eta > 0$ is another regularization parameter.
The detailed derivation is described in \ref{sec:class-prior-estimation}.

\subsection{Remarks: Instance-Level PU-MIL}
\label{sec:uu-classification}

PU-SKC is a {\it bag-level method}, namely, classifying bags directly (Eq.~\eqref{eq:bag-clf}) instead of aggregating instance-level classification results.
On the other hand, we can also consider an {\it instance-level method} to solve PU-MIL.

Assume that instances in negative bags are drawn from the instance-level negative conditional distribution, i.e.,
\begin{align}
  \bm{x} \stackrel{\mathrm{i.i.d.}}{\sim} p(\bm{x} | y = -1) \quad \text{for every $\bm{x} \in X^{\mathrm{N}}$}, \label{eq:negative-bag-distribution}
\end{align}
and instances in positive bags are drawn from the instance-level marginal distribution, i.e.,
\begin{align}
  \bm{x} \stackrel{\mathrm{i.i.d.}}{\sim} p(\bm{x}) = \theta p(\bm{x} | y = +1) + (1-\theta) p(\bm{x} | y = -1) \quad \text{for every $\bm{x} \in X^{\mathrm{P}}$}, \label{eq:positive-bag-distribution}
\end{align}
where $\theta := p(y = +1)$ is the instance-level class prior.
Since we assume the bag-level class prior $p(Y = +1) = \pi$, instances in unlabeled bags are drawn from the following distribution, i.e.,
\begin{align}
\bm{x} \stackrel{\mathrm{i.i.d.}}{\sim}
p'(\bm{x})
  &= \pi p(\bm{x}) + (1 - \pi) p(\bm{x} | y = -1) \nonumber \\
  &= \pi\theta p(\bm{x} | y = +1) + (1 - \pi\theta) p(\bm{x} | y = -1) \quad \text{for every $\bm{x} \in X^{\mathrm{U}}$}. \label{eq:unlabeled-bag-distribution}
\end{align}
From the instance-level perspective, both positive and unlabeled bags are unlabeled datasets, but the class proportions are different ($\theta$ for Eq.~\eqref{eq:positive-bag-distribution} and $\pi\theta$ for Eq.~\eqref{eq:unlabeled-bag-distribution}).
In fact, an (instance-level) binary classifier can be obtained from two distinct datasets with different class proportions~\cite{Plessis:2013}.

Assume that the test class priors are equal $q(y = +1) = q(y = -1) = 1/2$ and the test conditional density $q(\bm{x} | y)$ is equal to $p(\bm{x} | y)$.
We begin with the difference of the class posteriors:
\begin{align*}
  q(y = +1 | \bm{x}) - q(y = -1 | \bm{x})
  &= \frac{q(\bm{x} | y = +1)q(y = +1)}{q(\bm{x})} - \frac{q(\bm{x} | y = -1)q(y = -1)}{q(\bm{x})} \nonumber \\
  &= \frac{p(\bm{x} | y = +1)\frac{1}{2}}{q(\bm{x})} - \frac{p(\bm{x} | y = -1)\frac{1}{2}}{q(\bm{x})} \nonumber \\
  &= \frac{1}{2q(\bm{x})} (p(\bm{x} | y = +1) - p(\bm{x} | y = -1)) .
\end{align*}
Since $2q(\bm{x})$ is always positive, the classification criterion on the test distribution becomes
\begin{align*}
  d(\bm{x}) = \mathrm{sign}[ p(\bm{x} | y = +1) - p(\bm{x} | y = -1) ] .
\end{align*}
On the other hand,
\begin{align*}
  p(\bm{x}) - p'(\bm{x})
  &= (\theta - \pi\theta)p(\bm{x} | y = +1) - ((1 - \theta) - (1 - \pi\theta))p(\bm{x} | y = -1) \nonumber \\
  &= \theta(1 - \pi)[p(\bm{x} | y = +1) - p(\bm{x} | y = -1)] .
\end{align*}
Since $\theta(1 - \pi) > 0$, the classification criterion becomes\footnote{
In the original paper~\cite{Plessis:2013},
an unknown constant $C = \pm 1$
is multiplied in the right-hand side.
On the other hand, in our current setting, we know that the class priors are $\theta$ and $\pi\theta$ and this allows us to determine the sign of $C$.
}
\begin{align}
  d(\bm{x}) = \mathrm{sign}[p(\bm{x}) - p'(\bm{x})] . \label{eq:uu-classifier}
\end{align}
The point is, in order to obtain an instance-level classifier, all we have to do is to estimate the density difference $p(\bm{x}) - p'(\bm{x})$.
To this end, a method called {\it least-squares density difference (LSDD)} estimation has been proposed~\cite{Sugiyama:2012}.
A more advanced method to estimate the sign of the density difference $\mathrm{sign}[p(\bm{x}) - p'(\bm{x})]$ directly has also been proposed~\cite{Plessis:2013}, which is called {\it direct sign density difference (DSDD)} estimation.
These estimators are also compared as baselines in experiments.

The instance-level approach is useful when our goal is to determine all instance-level labels.
Otherwise, the bag-level approach is suitable because it is a direct approach to determine only bag-level labels,
which is referred to as Vapnik's principle~\cite{Vapnik:1998}:
\begin{displayquote}
  When solving a problem of interest, one should not solve a more general problem as an intermediate step.
\end{displayquote}
Here, knowing instance-level labels allows us to identify bag-level labels, but not vice versa.
In this sense, obtaining an instance-level classifier is regarded as solving an intermediate/general problem when our goal is to predict labels for bags.

% ----- theory ----- %

\section{Analysis of Generalization Error Bounds}
\label{sec:theory}

In this section, we show an upper bound of the generalization error (evaluated on a fixed classifier) for our proposed method.
Let $\mathscr{X} := \mathcal{P}(\mathbb{R}^d)\setminus\emptyset$ be the bag-level domain set
and
\begin{align}
  \mathcal{G} := \left\{
    g(X) = \bm{\alpha}^\top\bm{\phi}(X)
    \;\middle|\; \| \bm{\alpha} \| \le C_{\bm{\alpha}}, \sup_{X\in\mathscr{X}}\|\bm{\phi}(X)\| \le C_{\bm{\phi}}
  \right\} \label{eq:function-class}
\end{align}
be a given function class, where $\bm{\phi}$ is a vector of basis functions defined in Eq.~\eqref{eq:bag-basis}.
Note that $\mathcal{G}$ includes the classifier~\eqref{eq:bag-clf} as a special case\footnote{Let $\tilde{\bm{\alpha}}:=[\bm{\alpha}\;\beta]^\top$ and $\tilde{\bm{\phi}}(X):=[\bm{\phi}(X)\;1]^\top$ then it is included in $\mathcal{G}$.}.
Throughout this section, let $\ell: \mathbb{R} \rightarrow \mathbb{R}_+$ be the double hinge loss~\eqref{eq:dh-loss}, which is used in our experiments.
%Throughout this section, let $\ell: \mathbb{R} \rightarrow \mathbb{R}_+$ be a loss function which satisfy Eq.~\eqref{eq:linear-composite-loss}.
We denote the expected risk of a bag-level classifier $g: \mathscr{X} \rightarrow \mathbb{R}$ with respect to $\ell$ as
\begin{align*}
  \mathcal{R}(g) &:= \mathbb{E}_{p(X,Y)}[\ell(Yg(X))] \\
                 &= \pi^*\mathbb{E}_{p(X|Y=+1)}[-g(X)] + \mathbb{E}_{p(X)}[\ell(-g(X))],
\end{align*}
and the corresponding empirical risk as
\begin{align*}
  \hat{\mathcal{R}}(g) := -\frac{\pi^*}{N_\positive}\sum_{b=1}^{N_\positive} g(X_b^\positive)  + \frac{1}{N_\unlabeled}\sum_{b'=1}^{N_\unlabeled}\ell(-g(X_{b'}^\unlabeled)),
\end{align*}
where $\pi^* := p(Y = +1)$ is the true class prior of the positive class.

\begin{theorem}
  For a fixed $g \in \mathcal{G}$, and for any $\delta \in (0, 1)$, with probability at least $1 - \delta$,
  \begin{align}
    \mathcal{R}(g) - \hat{\mathcal{R}}(g) \le C_{\mathcal{G},\delta}\left(\frac{2\pi^*}{\sqrt{N_\positive}} + \frac{1}{\sqrt{N_\unlabeled}}\right), \label{eq:bound}
  \end{align}
  where $C_{\mathcal{G},\delta} > 0$ is a constant depending jointly on $\mathcal{G}$ and $\delta$.
\end{theorem}

The proof is in \ref{sec:proof}.
This theorem shows that the generalization error decreases with order $1/\sqrt{N_\positive}$ and $1/\sqrt{N_\unlabeled}$.
Thus, increasing the number of positive bags and the number of unlabeled bags both contributes to reducing the error.
Note that this order is optimal in a parametric setup \cite{Vapnik:1995}.
Furthermore, we can see from Eq.~\eqref{eq:bound} that the true class prior $\pi^*$ and $N_\positive$ are related in the generalization error bound, while $\pi^*$ and $N_\unlabeled$ are not. We will further investigate this issue through experiments in Section~\ref{sec:experiments}.

% ----- experiment ----- %

\section{Experiments}
\label{sec:experiments}

In this section, we experimentally compare the proposed method\footnote{
  Implementation is published at \url{https://github.com/levelfour/pumil}.
}
with the baselines (Sect.~\ref{sec:uu-classification}), and an existing method (see \ref{sec:pu-mil}) and give answers to the following research questions.

$\;$\\
Q1: Does the proposed method outperform the baseline and existing methods regardless of the true class prior? \\
Q2: Is the proposed method computationally efficient?

\subsection{Datasets}

We used standard MIL datasets: Musk and Corel\footnote{\url{http://www.cs.columbia.edu/~andrews/mil/datasets.html}}.
The details of these benchmark datasets are shown in Table \ref{tab:datasets}.
\begin{table}[t]
\begin{center}
\centering
\caption{Details of datasets: The last two rows indicate the numbers of instances per bag, which are the average with standard deviation.}
\begin{tabular}{ c||c|c|c|c|c } \hline
    Number of & Musk1 & Musk2 & Elephant & Fox & Tiger \\ \hline
    features & 166 & 166 & 230 & 230 & 230 \\
    positive bags & 47 & 39 & 100 & 100 & 100 \\
    negative bags & 45 & 63 & 100 & 100 & 100 \\
    positive instances & 2.3 (2.6) & 10.0 (26.1) & 3.8 (4.2) & 3.2 (3.6) & 2.7 (3.1) \\
    negative instances & 2.9 (6.9) & 54.7 (176.0) & 3.2 (3.6) & 3.4 (3.8) & 3.4 (3.8) \\ \hline
\end{tabular}
\label{tab:datasets}
\end{center}
\end{table}

Since these datasets are too small to evaluate PU methods, we augmented them to increase the number of bags.
Specifically, bags chosen randomly from the original datasets were duplicated
and then Gaussian noise with mean zero and variance $0.01$ was added to each dimension.
In this way, we increased the number of samples in the Musk datasets (Musk1 and Musk2) 10 times and the Corel datasets (Elephant, Fox, and Tiger) 5 times.
After this augmentation process, we generated a training set (including labeled positive bags and unlabeled bags) and a test set.
This generation process is described in Algorithm~\ref{alg:benchmark-set-generation} (we set $L = 20, U = 180$, and $T = 200$).

\textit{Remark:} This dataset processing is needed. The reasons are as follows.
\begin{enumerate}
  \item
  We assume that the training distribution and test distribution are same,
  which means that the class priors are same, too.
  Thus Algorithm~\ref{alg:benchmark-set-generation} is needed to maintain the class priors to be same among both (unlabeled) training and test datasets.

  \item
  If Algorithm~\ref{alg:benchmark-set-generation} is applied, it is hard to obtain an enough number of negative bag samples under extremely low class priors,
  while maintaining the class priors to be same.
  Thus the augmentation process is needed.
\end{enumerate}

\begin{algorithm}[t]
\caption{Generation of Training / Test Sets for Benchmark MIL Datasets} \label{alg:benchmark-set-generation}
\begin{algorithmic}[1]
\Require $\mathcal{D}_\positive$: original positive bags, $\mathcal{D}_\negative$: original negative bags, $\pi$: true bag-level class prior,
$L$: \#\{labeled positive bags\}, $U$: \#\{unlabeled bags\}, $T$: \#\{test bags\}
\State $\mathcal{D}_\mathrm{L} \subset \mathcal{D}_\positive$ \Comment{$\lvert\mathcal{D}_\mathrm{L}\rvert = L$}
\State $\mathcal{D}_\positive := \mathcal{D}_\positive \setminus \mathcal{D}_\mathrm{L}$
\State $N_\unlabeled^\positive \sim B(U + T, \pi)$
\State $N_\unlabeled^\negative := U + T - N_\unlabeled^\positive$
\State $\mathcal{D}_\unlabeled := \mathcal{D}_\positive'(\subset\mathcal{D}_\positive) \cup \mathcal{D}_\negative'(\subset\mathcal{D}_\negative)$
\Comment{$\lvert\mathcal{D}_\positive'\rvert = N_\unlabeled^\positive, \lvert\mathcal{D}_\negative'\rvert = N_\unlabeled^\negative$}
\State $\mathcal{D}_\unlabeled' \subset \mathcal{D}_\unlabeled$ \Comment{$\lvert\mathcal{D}_\unlabeled'\rvert = T$}
\State $\mathcal{D}_\unlabeled := \mathcal{D}_\unlabeled \setminus \mathcal{D}_\unlabeled'$
\Ensure $\mathcal{D}_\mathrm{L} \cup \mathcal{D}_\unlabeled$: training set, $\mathcal{D}_\unlabeled'$: test set
\end{algorithmic}
\end{algorithm}

\subsection{Methods}

We compared the following methods:
\begin{itemize}
  \item {\bf Positive-Unlabeled Set Kernel Classifier (PU-SKC, the proposed method)}:
  Hyperparameters (the degree parameter $\rho$ in the polynomial kernel~\eqref{eq:set-kernel} and the regularization parameter $\lambda$ in the objective function~\eqref{eq:pu-skc-obj}) were selected via 5-fold cross-validation from $\rho \in [1, 2, 3]$ and $\lambda \in [10^0, 10^{-3}, 10^{-6}]$.
  Values minimizing the PU risk~\eqref{eq:PU-convex-obj} with the zero-one loss were chosen to be optimal.

  \item {\bf Least-Squares Density Difference (LSDD)}:
  Estimate the density difference $p(\bm{x}) - p'(\bm{x})$ to obtain~\eqref{eq:uu-classifier} using the least-squares method~\cite{Sugiyama:2012}.
  The bag classifier can be obtained as $g(X) = \max_{\bm{x} \in X} d(\bm{x})$.
  Hyperparameters (the width of the Gaussian kernel $s$ and the regularization parameter $\lambda$) were selected via 5-fold cross-validation from $s \in [10^{-2}, 10^{-4}, 10^{-6}]$ and $\lambda \in [10^0, 10^{-3}, 10^{-6}]$.

  \item {\bf Direct Sign of Density Difference (DSDD)}:
  Estimate the sign of the density difference~\eqref{eq:uu-classifier} directly~\cite{Plessis:2013}.
  The bag classifier can be obtained in the same way as LSDD.
  Hyperparameters (the width of the Gaussian kernel $s$ and the regularization parameter $\lambda$) were selected via 5-fold cross-validation from $s \in [10^{-2}, 10^{-4}, 10^{-6}]$ and $\lambda \in [10^0, 10^{-3}, 10^{-6}]$.

  \item {\bf puMIL}:
  An SVM-based approach~\cite{Wu:2014}. The detail is described in \ref{sec:pu-mil}.
\end{itemize}

\subsection{Results}

Here we show the experimental results and give answers to the research questions.

\subsubsection{Classification Performances}

\begin{table}[!ht]
\caption{Each result is the average with standard deviation of the classification accuracy over 20 trials. Bold faces represent the best methods under each setting (tested by the one-sided 5\% t-test).}
\centering
\footnotesize
\begin{tabular}{lc|rrrr} \hline
% header
dataset & $\pi$ & PU-SKC & LSDD~\cite{Sugiyama:2012} & DSDD~\cite{Plessis:2013} & puMIL~\cite{Wu:2014} \\ \hline
Musk1      & 0.1 &         0.865 (0.046)  & \textbf{0.928 (0.029)} & \textbf{0.931 (0.026)} &         0.757 (0.065)  \\
           & 0.2 &         0.844 (0.038)  &         0.707 (0.289)  & \textbf{0.876 (0.037)} &         0.733 (0.070)  \\
           & 0.3 & \textbf{0.818 (0.041)} &         0.622 (0.208)  &         0.778 (0.045)  &         0.717 (0.063)  \\
           & 0.4 & \textbf{0.776 (0.057)} &         0.618 (0.148)  &         0.708 (0.041)  &         0.699 (0.050)  \\
           & 0.5 & \textbf{0.763 (0.050)} &         0.553 (0.072)  &         0.597 (0.047)  &         0.665 (0.081)  \\
           & 0.6 & \textbf{0.735 (0.055)} &         0.522 (0.068)  &         0.505 (0.051)  &         0.649 (0.069)  \\
           & 0.7 & \textbf{0.737 (0.044)} &         0.538 (0.161)  &         0.392 (0.061)  &         0.606 (0.075)  \\\hline
Musk2      & 0.1 &         0.810 (0.060)  &         0.702 (0.303)  & \textbf{0.840 (0.040)} &         0.688 (0.098)  \\
           & 0.2 & \textbf{0.802 (0.053)} &         0.691 (0.239)  & \textbf{0.789 (0.036)} &         0.730 (0.063)  \\
           & 0.3 & \textbf{0.801 (0.051)} &         0.621 (0.173)  &         0.720 (0.050)  &         0.732 (0.073)  \\
           & 0.4 & \textbf{0.724 (0.063)} &         0.590 (0.111)  &         0.621 (0.048)  & \textbf{0.704 (0.068)} \\
           & 0.5 & \textbf{0.742 (0.050)} &         0.522 (0.054)  &         0.537 (0.055)  &         0.654 (0.092)  \\
           & 0.6 & \textbf{0.706 (0.059)} &         0.499 (0.086)  &         0.466 (0.053)  &         0.637 (0.086)  \\
           & 0.7 & \textbf{0.726 (0.055)} &         0.508 (0.159)  &         0.364 (0.063)  &         0.599 (0.072)  \\\hline
Elephant   & 0.1 & \textbf{0.845 (0.044)} &         0.734 (0.153)  &         0.747 (0.041)  &         0.722 (0.084)  \\
           & 0.2 & \textbf{0.783 (0.062)} &         0.671 (0.166)  &         0.715 (0.039)  &         0.686 (0.075)  \\
           & 0.3 & \textbf{0.746 (0.062)} &         0.652 (0.092)  &         0.685 (0.038)  &         0.698 (0.072)  \\
           & 0.4 & \textbf{0.701 (0.050)} &         0.573 (0.088)  &         0.608 (0.039)  &         0.642 (0.051)  \\
           & 0.5 &         0.607 (0.058)  &         0.552 (0.052)  &         0.575 (0.050)  & \textbf{0.667 (0.070)} \\
           & 0.6 &         0.528 (0.087)  &         0.520 (0.062)  &         0.503 (0.031)  & \textbf{0.614 (0.063)} \\
           & 0.7 &         0.421 (0.059)  & \textbf{0.602 (0.141)} &         0.453 (0.037)  & \textbf{0.597 (0.066)} \\\hline
Fox        & 0.1 & \textbf{0.840 (0.053)} &         0.634 (0.181)  &         0.717 (0.037)  &         0.561 (0.062)  \\
           & 0.2 & \textbf{0.754 (0.048)} &         0.689 (0.042)  &         0.669 (0.041)  &         0.575 (0.058)  \\
           & 0.3 & \textbf{0.689 (0.054)} &         0.576 (0.130)  &         0.615 (0.045)  &         0.544 (0.045)  \\
           & 0.4 & \textbf{0.613 (0.061)} &         0.542 (0.097)  & \textbf{0.588 (0.041)} &         0.552 (0.053)  \\
           & 0.5 & \textbf{0.538 (0.042)} & \textbf{0.527 (0.046)} & \textbf{0.545 (0.042)} & \textbf{0.547 (0.074)} \\
           & 0.6 &         0.468 (0.057)  & \textbf{0.538 (0.071)} &         0.464 (0.053)  & \textbf{0.550 (0.055)} \\
           & 0.7 &         0.430 (0.055)  & \textbf{0.559 (0.151)} &         0.430 (0.046)  & \textbf{0.536 (0.075)} \\\hline
Tiger      & 0.1 & \textbf{0.839 (0.044)} &         0.689 (0.136)  &         0.730 (0.047)  &         0.706 (0.073)  \\
           & 0.2 & \textbf{0.783 (0.035)} &         0.695 (0.032)  &         0.707 (0.031)  &         0.680 (0.070)  \\
           & 0.3 & \textbf{0.729 (0.043)} &         0.636 (0.047)  &         0.651 (0.052)  &         0.659 (0.056)  \\
           & 0.4 & \textbf{0.672 (0.043)} &         0.596 (0.042)  &         0.606 (0.036)  & \textbf{0.668 (0.048)} \\
           & 0.5 & \textbf{0.584 (0.043)} &         0.566 (0.044)  &         0.540 (0.046)  & \textbf{0.609 (0.070)} \\
           & 0.6 &         0.508 (0.048)  &         0.530 (0.066)  &         0.523 (0.032)  & \textbf{0.608 (0.080)} \\
           & 0.7 &         0.410 (0.059)  &         0.490 (0.080)  &         0.463 (0.034)  & \textbf{0.585 (0.086)} \\\hline
\end{tabular}
\label{tab:pumil-auc}
\end{table}

Table \ref{tab:pumil-auc} shows averages with standard deviations of the classification accuracy over $20$ trials under each class prior.
Bold faces represent the best methods under each class prior.
This was tested by the one-sided t-test with 5\% significance level
(first the best method was chosen, then other methods were checked whether they are comparable or not by the one-sided t-test).
As it can be seen from Table \ref{tab:pumil-auc}, PU-SKC outperforms the existing method puMIL~\cite{Wu:2014} under various class priors.
Note that the true class prior in Table \ref{tab:pumil-auc} means the predefined value for dataset generation (see Algorithm \ref{alg:benchmark-set-generation}),
not an estimated class prior during the learning process (see Sect.~\ref{sec:bag-class-prior-estimation}).

\begin{figure}[t]
  \caption{Classification performances of each dataset under different number of positive bags. The number of unlabeled bags are fixed to 180. The true class priors are set to $0.7$ among all settings.}
  \label{fig:np-change}
  \begin{center}
    \includegraphics[width=180px]{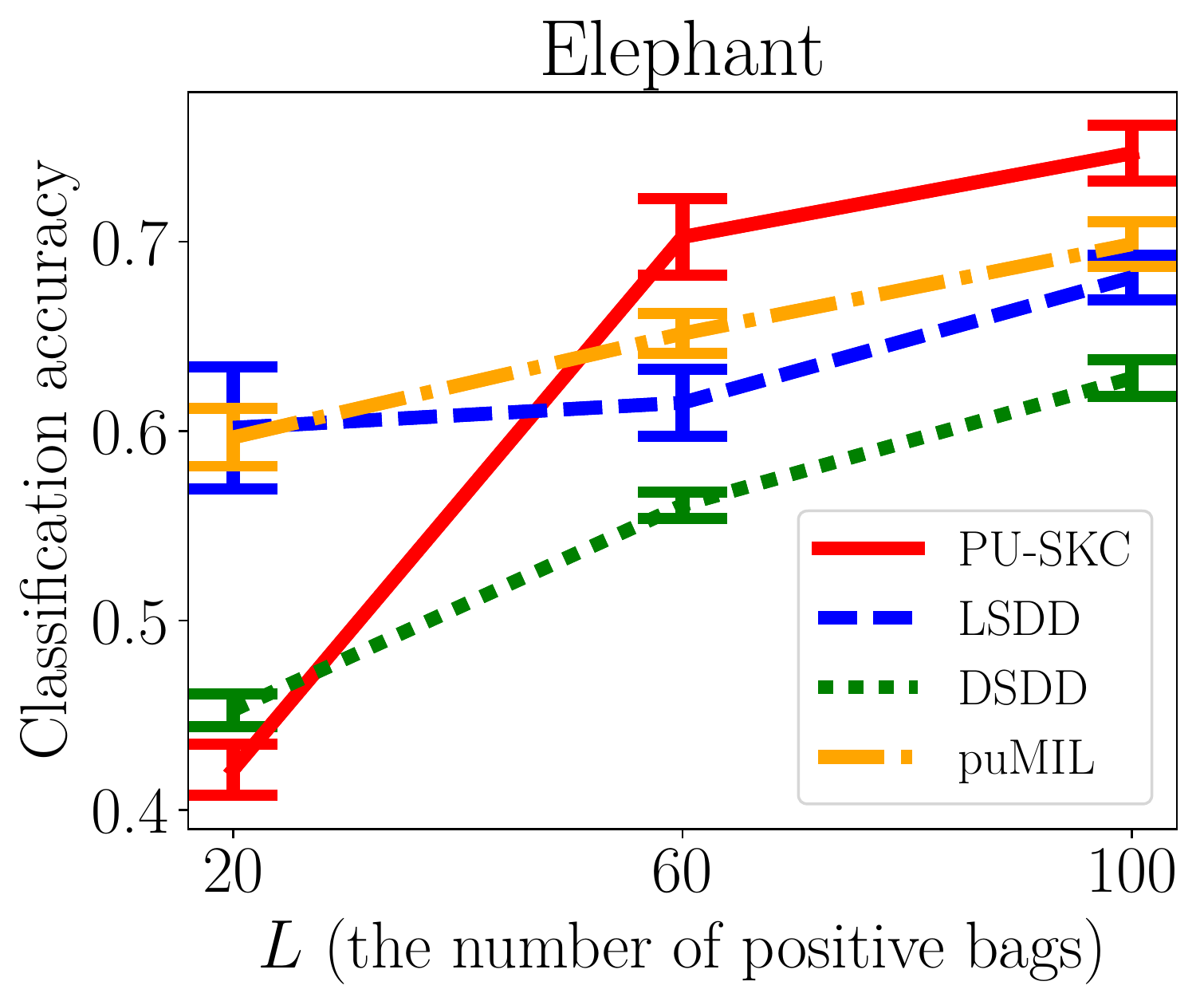}
    \includegraphics[width=180px]{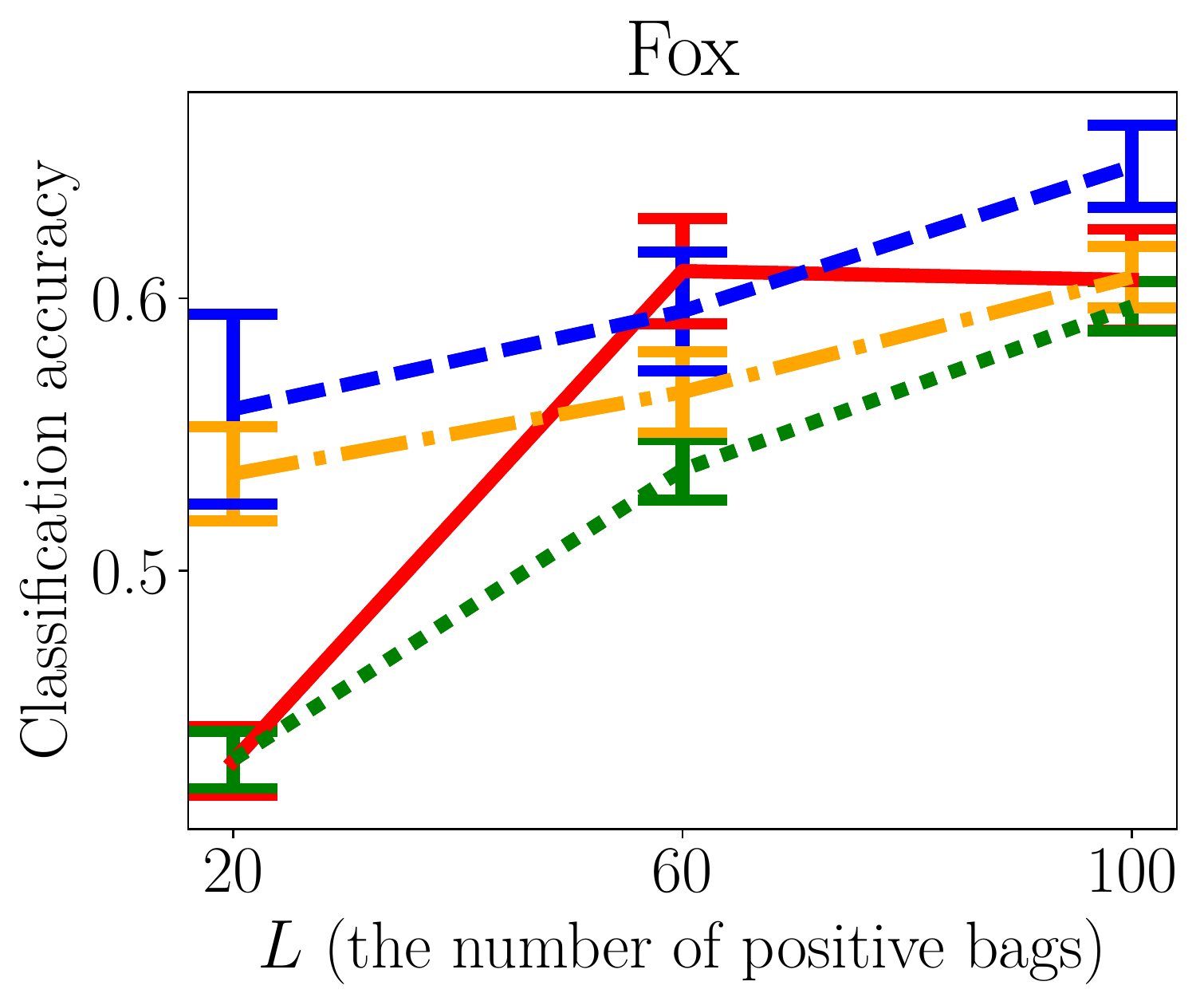}
    \includegraphics[width=180px]{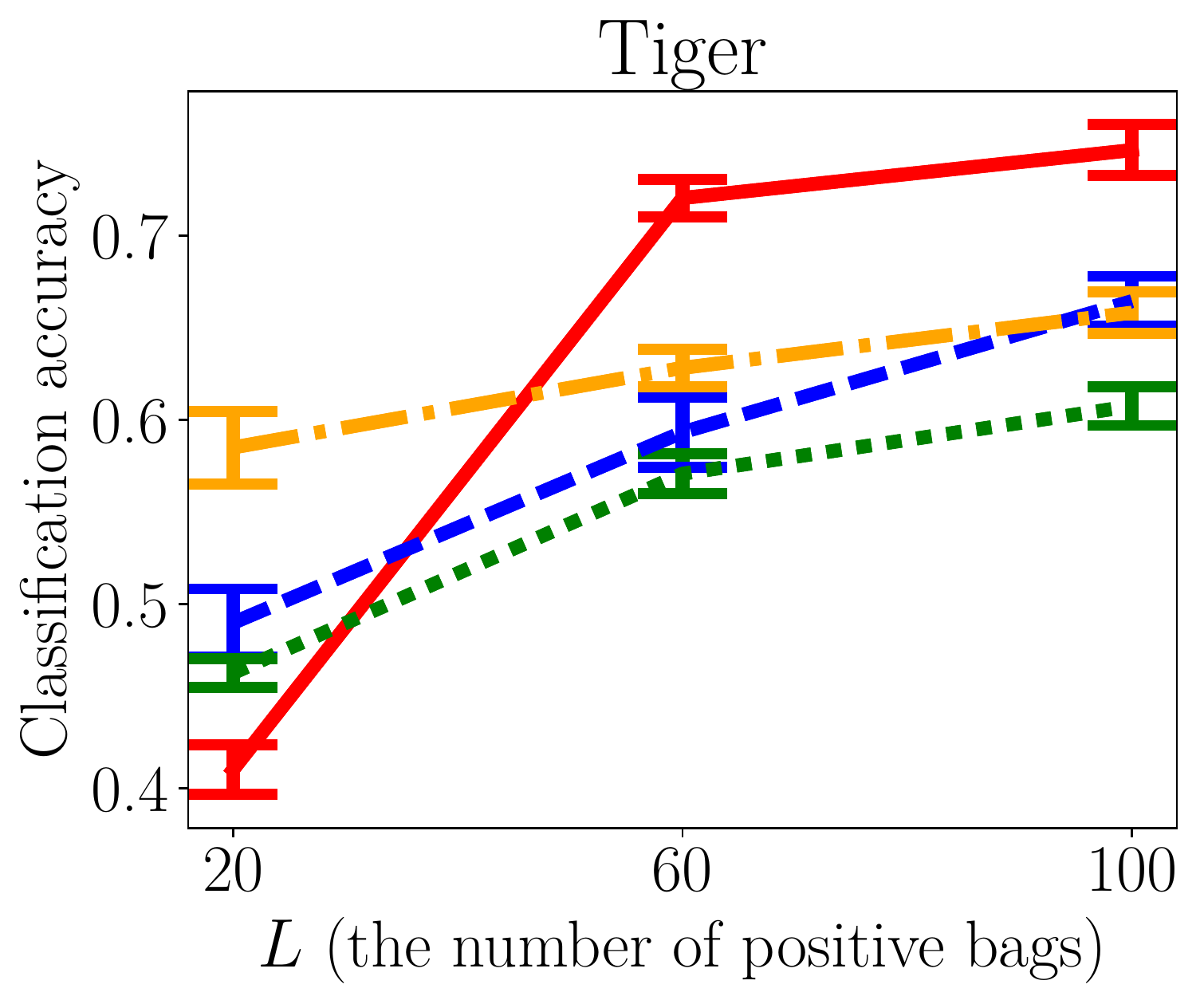}
  \end{center}
\end{figure}

Overall, the performance of the proposed method decreases as the true class prior becomes higher.
This can be confirmed from Eq.~\eqref{eq:bound}:
since $2\pi^*/\sqrt{N_\positive}$ dominates the upper bound given sufficiently large $N_\unlabeled$,
more positive bags are needed to achieve accurate classification performance compared with a small class prior case.
In practice, we could address this issue by collecting more positive bags.
Figure~\ref{fig:np-change} shows classification performances under different number of positive bags.
Classification performances are improved as the number of positive bags increases.\\

\textbf{A1: The proposed method tends to outperform the baseline and existing methods under various class priors.}\\

\subsubsection{Computation Time}

\begin{figure}[t]
    \caption{
      Average execution time: each result is the average execution time of 20 trials under bag-level class prior $\pi = 0.1$.
      PU-SKC is executed on the given hyperparameter.
    }
    \centering\includegraphics[width=10cm]{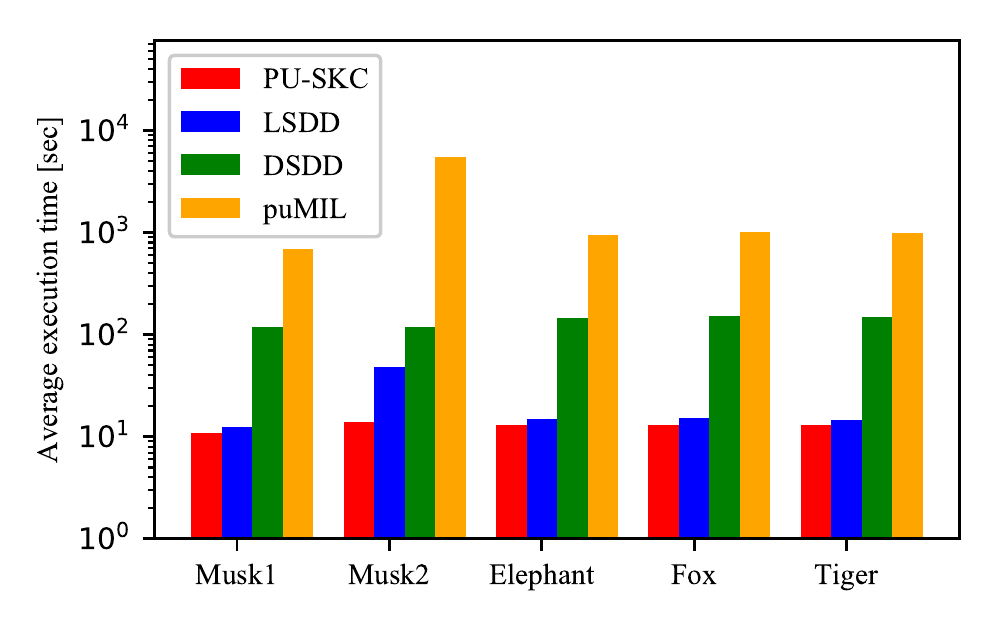}
    \label{fig:elapsed-time}
\end{figure}

Next, we compared the execution time between the proposed method and the baseline and existing methods.
The result is shown in Figure~\ref{fig:elapsed-time}.
This result shows that PU-SKC is much more computationally efficient than the baseline and existing methods.
Note that the execution time in other class prior values ($\pi = 0.2, 0.3, \dots, 0.7$) is almost the same as the one shown in Figure~\ref{fig:elapsed-time}
because both the class prior estimation algorithm shown in Sec~\ref{sec:bag-class-prior-estimation} and the PU-SKC optimization problem~\eqref{eq:puskc-qp} are non-iterative methods and their computation complexities do not depend on the value of $\pi$ or its estimated value.\\

\textbf{A2: The proposed method is much more computationally efficient than the baseline and existing methods.}\\

%From Figure \ref{fig:elapsed-time}, we can see the computation time of puMIL on Musk2 is about 10 times more than that on the other datasets.
%This is because Musk2 has more instances in each bag than the other datasets (see Table \ref{tab:datasets}), and the computational cost of the estimator \eqref{eq:WKDE} is $\mathcal{O}(N)$, where $N$ denotes the number of possibly negative instances.
%On the other hand, PU-SKC uses the minimax kernel \eqref{eq:set-kernel}, whose computational cost does not depend on the number of instances.

% ----- conclusion ----- %

\section{Conclusion}
\label{sec:conclusion}

In this work, we considered a multiple instance learning problem when only positive bags and unlabeled bags are available, which does not require all training bags to be labeled.
We proposed a convex method, PU-SKC, to solve PU multiple instance classification.
This method is based on the convex formulation of PU classification~\cite{Plessis:2015} and the set kernel~\cite{Gartner:2002}.
PU-SKC performed better than the existing PU multiple instance classification method~\cite{Wu:2014} on benchmark MIL datasets.
Furthermore, we confirmed that the proposed method was much more computationally efficient than the baseline and existing methods through the experiment.

% ----- acknowledgement ----- %
\section*{Acknowledgement}
TS was supported by KAKENHI 15J09111.

% ----- references ----- %
\section*{Reference}
\bibliographystyle{elsarticle-num}
\bibliography{ref}

\newpage

% ----- appendix ----- %

\appendix
\section{Instance-Level Class Prior Estimation}
\label{sec:class-prior-estimation}

In the research by du Plessis and Sugiyama~\cite{Plessis:2014b}, a class prior estimation method from positive and unlabeled data by partial matching was proposed.
This method estimates the class prior $\pi$ by minimizing the Pearson (PE) divergence from positive data distribution $\pi p(\bm{x} | y = 1)$ to unlabeled data distribution $p(\bm{x})$:
\begin{align*}
\pi^*
& = \argmin_\pi \textsf{PE}[\pi p(\bm{x}|y=1) | p(\bm{x})] \\
& = \argmin_\pi \frac{1}{2}\int\left(\frac{\pi p(\bm{x}|y=1)}{p(\bm{x})} - 1\right)^2 p(\bm{x})\D\bm{x} .
\end{align*}
It is still possible to estimate $p(\bm{x}|y=1)$ from positive samples and $p(\bm{x})$ from unlabeled samples using, e.g., the kernel density estimation,
but du Plessis and Sugiyama~\cite{Plessis:2014b} empirically showed that such a naive approach often does not produce a good estimator of $\pi$.
A better approach is to lower bound the PE divergence and directly maximize it:
\begin{align}
\textsf{PE}[\pi p(\bm{x}|y=1) | p(\bm{x})]
\ge & \; \pi \int r(\bm{x})p(\bm{x}|y=1)\D\bm{x} \nonumber \\
    & - \frac{1}{2} \int r(\bm{x})^2 p(\bm{x})\D\bm{x}
      - \pi + \frac{1}{2} \label{eq:PE-lower-bound} ,
\end{align}
where $r(\bm{x})$ is an arbitrary function.
In practice, a linear-in-parameter model based on Gaussian kernel basis functions is used to estimate $r$:
\[
\widehat{r}(\bm{x}) = \bm{\alpha}^\top\bm{\phi}(\bm{x}) .
\]
Our goal is to find the tightest lower bound by maximizing the right-hand side of (\ref{eq:PE-lower-bound}) at first with respect to $r$ and then minimize it with respect to $\pi$.
The former maximization problem with the $l_2$ regularizer can be written as
\begin{align}
\widehat{\bm{\alpha}}
&= \argmax_{\bm{\alpha}} \left[ \pi\bm{\alpha}^\top\bm{h} - \frac{1}{2}\bm{\alpha}^\top H\bm{\alpha} - \pi + \frac{1}{2} - \frac{\lambda}{2}\bm{\alpha}^\top\bm{\alpha} \right] \nonumber \\
&= \argmax_{\bm{\alpha}} \left[ \pi\bm{\alpha}^\top\bm{h} - \frac{1}{2}\bm{\alpha}^\top H\bm{\alpha} - \frac{\lambda}{2}\bm{\alpha}^\top\bm{\alpha} \right] , \label{eq:PE-lb} \\
\mbox{where} \quad
H     &= \int\bm{\phi}(\bm{x})\bm{\phi}(\bm{x})^\top p(\bm{x})\D\bm{x}, \quad
\bm{h} = \int\bm{\phi}(\bm{x})p(\bm{x}|y=1)\D\bm{x} , \nonumber
\end{align}
and $\lambda \ge 0$ denotes the regularization parameter.
In practice, $H$ and $\bm{h}$ are estimated by the sample averages:
\[
\widehat{H} = \frac{1}{N_\positive} \sum_{i=1}^{N_\positive} \bm{\phi}(\bm{x}^\positive_i)\bm{\phi}(\bm{x}^\positive_i)^\top, \quad
\widehat{\bm{h}} = \frac{1}{N_\unlabeled} \sum_{j=1}^{N_\unlabeled} \bm{\phi}(\bm{x}^\unlabeled_j) .
\]
Using these estimators, Eq.~\eqref{eq:PE-lb} can be reformulated as follows:
\[
\widehat{\bm{\alpha}} = \argmax_{\bm{\alpha}} \left[ \pi\bm{\alpha}^\top\widehat{\bm{h}} - \frac{1}{2}\bm{\alpha}^\top\widehat{H}\bm{\alpha} - \frac{\lambda}{2}\bm{\alpha}^\top\bm{\alpha} \right] .
\]
An analytical solution to the above problem can be obtained as follows:
\[
\widehat{\bm{\alpha}} = \pi\widehat{G}^{-1}\widehat{\bm{h}}, \quad \widehat{G} = \widehat{H} + \lambda I ,
\]
where $I$ denotes the identity matrix.
This leads to the following PE divergence estimator:
\[
\widehat{\textsf{PE}} = \pi^2\widehat{\bm{h}}^\top\widehat{G}^{-1}\widehat{\bm{h}} - \pi^2\frac{1}{2}\widehat{\bm{h}}^\top\widehat{G}^{-1}\widehat{H}\widehat{G}^{-1}\widehat{\bm{h}} - \pi + \frac{1}{2}.
\]
Then the analytical minimizer of $\widehat{\textsf{PE}}$ can be obtained as
\[
\widehat{\pi}  = \left[
2\widehat{\bm{h}}^\top\widehat{G}^{-1}\widehat{\bm{h}} - \widehat{\bm{h}}^\top\widehat{G}^{-1}\widehat{H}\widehat{G}^{-1}\widehat{\bm{h}}
\right]^{-1}.
\]

\section{Multiple Instance Learning from Positive and Unlabeled Bags}
\label{sec:pu-mil}

Wu et al.~\cite{Wu:2014} proposed an instance-level method for the PU-MIL problem (called puMIL).
As discussed in the previous work~\cite{Andrews:2002,Li:2009:ECML}, the key instance (the most positive instance) in each bag is important in MIL.
This is also the case in PU-MIL, but the problem is that we cannot tell the labels of unlabeled bags.
To address this issue, Wu et al.~\cite{Wu:2014} introduced the bag confidence, which describes how much confident the given unlabeled bag is negative.
Let $\nu_b \in [0, 1]$ be the confidence of the $b$-th bag.
Then the learning problem of an SVM classifier $g(X) = \bm{w}^\top\widetilde{\bm{x}} \quad (\widetilde{\bm{x}} \in X)$ under the given $\bm{\nu}$ is formulated as
\begin{align}
\min_{\bm{w} \in \mathbb{R}^d, \{\widetilde{\bm{x}}_b\}_{b=1}^{N_\positive + N_\unlabeled}} \quad
& \frac{1}{2}||\bm{w}||^2
  + C \sum_{b=1}^{N_\positive + N_\unlabeled} \max(0, 1 - \nu_b \widetilde{Y}_b \bm{w}^\top \widetilde{\bm{x}}_b) ,
\label{eq:puMIL}
\end{align}
where $\widetilde{Y}_b$ is the estimated bag label ($\widetilde{Y}_b = +1$ for a positive bag and $\widetilde{Y}_b = -1$ for a possibly negative bag coming from the set of unlabeled bags)
and $\widetilde{\bm{x}}_b$ is the key instance in the bag $X_b$.
$\nu_b$ weighs the contribution of $\widetilde{\bm{x}}_b$ according to its confidence.
For positive bags, the confidence is simply set as $\nu_b = 1$.
The overall training method is summarized as follows.
\begin{enumerate}
  \item Initialize $L$ distinct bag confidences $\bm{\nu}^{(1)}, \ldots, \bm{\nu}^{(L)}$, where $L$ is the number of the bag confidences. \label{enum:init}
  \item Extract $N_\positive$ {\it possibly} negative bags from unlabeled bags. \label{enum:extract}
  \item Make a positive margin pool (PMP), which consists of the key instances from positive bags and possibly negative bags. \label{enum:PMP}
  \item Solve~\eqref{eq:puMIL} with using the PMP to obtain an SVM classifier, which is evaluated by the F-measure. \label{enum:clf}
\end{enumerate}

In Step \ref{enum:extract}, unlabeled bags with the highest confidences are extracted and regarded as possibly negative.
Bag confidences are initialized randomly and updated via a genetic algorithm (see Step \ref{enum:clf} and~\eqref{eq:mutation}).

In Step \ref{enum:PMP}, first the generative distribution of negative instances is estimated by a weighted kernel density estimator~\cite{Fu:2011}:
\begin{align}
\widehat{p}(\bm{x}|y = -1) = \dfrac{1}{\widetilde{N}_\negative} \sum_{b=1}^{\widetilde{N}_\negative} \sum_{\bm{x}' \in \widetilde{X}_b^\negative} k(\bm{x}, \nu_b \bm{x}') , \label{eq:WKDE}
\end{align}
where $\widetilde{N}_\negative$ is the number of all instances included in the extracted bags in Step \ref{enum:extract} and $k$ denotes a kernel function and $\widetilde{X}_b^\negative$ is the $b$-th possibly negative bag.
Using the estimator~\eqref{eq:WKDE}, we can obtain a PMP by extracting the key instances from bags:
\[
\widetilde{\bm{x}}_b = \argmin_{\bm{x}_{b} \in X_b} \widehat{p}(\bm{x}_{b} | y = -1) .
\]

After building the PMP, an SVM classifier can be obtained in Step \ref{enum:clf}.
We evaluate the given bag confidences $\bm{\nu}$ by calculating the F-measures\footnote{
    True bag labels are needed to calculate the F-measure, but those of unlabeled bags are unavailable.
    This point was not discussed in Wu et al.~\cite{Wu:2014}, so we assume that possibly negative bags have negative labels and the other bags in the unlabeled set have positive labels when we calculate the F-measure in our experiments.
}.
Here we have $L$ bag confidences, and we update them by a genetic algorithm.
First, the bag confidence with the highest F-measure (denoted by $\widehat{\bm{\nu}}$) is cloned to replace the other confidences under the fixed rate $c \in [0, 1]$.
This process, called the mutation, is caused by
\begin{align}
\bm{v}_{l} = \bm{\nu}_{l} + r (1 - f_l) (\widehat{\bm{\nu}} - \bm{\nu}_{l}) , \label{eq:mutation}
\end{align}
where $r$ is drawn from the standard normal distribution, $\bm{\nu}_{l}$ is the $l$-th bag confidence, and $f_l$ is the F-measure obtained from $\bm{\nu}_{l}$.
$\bm{\nu}_{l}$ is to be replaced for $\bm{v}_{l}$ if $\bm{v}_{l}$ produces a better result.
We iterate Steps \ref{enum:extract} -- \ref{enum:clf} until the best bag confidence $\widehat{\bm{\nu}}$ converges or the number of iterations reaches the predefined limit.

After the above training steps, the best bag confidence $\widehat{\bm{\nu}}$ is to be obtained.
We use $\widehat{\bm{\nu}}$ for the initial bag confidence and iterate Steps \ref{enum:extract} -- \ref{enum:clf} again to obtain a classifier for test prediction\footnote{
    In test prediction, test bag confidences are not defined.
    This point was also not discussed in Wu et al.~\cite{Wu:2014}, so we set all test bag confidences to 1 in our experiments.
}.
However, the process to extract possibly negative bags from unlabeled bags based on the bag confidences (Step \ref{enum:extract}) is not shown to converge to the optimal solution in the previous work, while this method experimentally works well.
%In the following section, we use the PU risk minimization framework \eqref{eq:PU-convex-obj} to formulate PU-MIL.

\section{Proof of Theorem 1}
\label{sec:proof}

\begin{proof}[\unskip\nopunct]
%  Let $C_\ell := \sup_{X\in\mathscr{X},y\in\{\pm 1\}}|\ell(y,g(X))|$
%  and $\rho > 0$ be the Lipschitz constant of $\ell$ on the restricted domain $\{yt \mid y \in \{\pm 1\}, \exists X\in\mathscr{X}. \; t=g(X)\} \subset \mathbb{R}$.
  By McDiarmid's inequality~\cite{McDiarmid:1989}, we obtain
  \begin{align*}
    \mathbb{P}\left( \mathbb{E}_{p(X|Y=+1)}[-g(X)] - \frac{1}{N_\positive}\sum_{b=1}^{N_\positive}(-g(X_b^{\positive})) \ge \epsilon \right)
    \le \exp\left(-\frac{2\epsilon^2}{N_\positive(2C_{\bm{\alpha}}C_{\bm{\phi}}/N_\positive)^2}\right)
    ,
  \end{align*}
  because $|-g(\cdot)| \le C_{\bm{\alpha}}C_{\bm{\phi}}$ by the Cauchy-Schwartz inequality (see the definition~\eqref{eq:function-class}).
  This is equivalent to the following inequality with probability at least $1 - \frac{\delta}{2}$:
  \begin{align*}
    \mathbb{E}_{p(X|Y=+1)}[-g(X)] - \frac{1}{N_\positive}\sum_{b=1}^{N_\positive}(-g(X_b^{\positive}))
    \le \sqrt{\frac{2C_{\bm{\alpha}}^2C_{\bm{\phi}}^2\log\frac{2}{\delta}}{N_\positive}}.
  \end{align*}
  Similarly, with probability at least $1 - \frac{\delta}{2}$,
  \begin{align*}
    \mathbb{E}_{p(X)}[\ell(-g(X))] - \frac{1}{N_\unlabeled}\sum_{b'=1}^{N_\unlabeled}\ell(-g(X_{b'}^\unlabeled))
    \le \sqrt{\frac{C_{\bm{\alpha}}^2C_{\bm{\phi}}^2\log\frac{2}{\delta}}{2N_\unlabeled}},
  \end{align*}
  noticing that $\sup_{X,X'\in\mathscr{X}}|\ell(-g(X)) - \ell(-g(X'))| \le C_{\bm{\alpha}}C_{\bm{\phi}} - 0 = C_{\bm{\alpha}}C_{\bm{\phi}}$ ($\because$~the Lipschitz constant of $\ell$ is $1$) when applying McDiarmid's inequality.

  Now let us move on to the generalization error bound.
  \begin{align*}
    \mathcal{R}(g) &- \hat{\mathcal{R}}(g)
    \\
    &= \pi^*\left\{\mathbb{E}_{p(X|Y=+1)}[-g(X)] - \frac{1}{N_\positive}\sum_{b=1}^{N_\positive}(-g(X_b^\positive))\right\} \\
    & \qquad + \left\{\mathbb{E}_{p(X)}[\ell(-g(X))] - \frac{1}{N_\unlabeled}\sum_{b'=1}^{N_\unlabeled}\ell(-g(X_{b'}^\unlabeled))\right\}
    \\
    &\le \sqrt{\frac{C_{\bm{\alpha}}^2C_{\bm{\phi}}^2\log\frac{2}{\delta}}{2}}\left(\frac{2\pi^*}{\sqrt{N_\positive}} + \frac{1}{\sqrt{N_\unlabeled}}\right),
  \end{align*}
  with probability at least $1 - \delta$,
  which concludes Eq.~\eqref{eq:bound} with $C_{\mathcal{G},\delta} = \sqrt{C_{\bm{\alpha}}^2C_{\bm{\phi}}^2\log\frac{2}{\delta}/2}$.
\end{proof}

\end{document}